%% file: submission.tex
\documentclass[10pt,journal,compsoc]{IEEEtran}
\ifCLASSOPTIONcompsoc
  \usepackage[nocompress]{cite}
\else
  \usepackage{cite}
\fi
\ifCLASSINFOpdf
\else
\fi
\usepackage{epsfig}
\usepackage{graphicx}
\usepackage{amsmath}
\usepackage{amssymb}
\usepackage{subfiles}
\usepackage{multirow} 
\newcommand{\ra}[1]{\renewcommand{\arraystretch}{#1}}
\usepackage{booktabs}
\usepackage[ruled]{algorithm2e}
\usepackage{xcolor,colortbl}
\usepackage{cite}
\usepackage{float}
\usepackage{booktabs}
\usepackage{rotating}
\usepackage{mmstyle}

\usepackage{enumitem}
\usepackage{transparent}

\usepackage{makecell}
\usepackage[pagebackref=true,breaklinks=true,colorlinks,bookmarks=false]{hyperref}
\definecolor{myred}{RGB}{225,97,78}

\def\Datasetname{Bamboo}
\def\LN{BambooTX}

\def\UnlabelPool{\mathcal{P}^{U}}
\def\LastUnlabelPool{\mathcal{P}^{U}(r-1)}
\def\CurUnlabelPool{\mathcal{P}^{U}(r)}

\def\LabelPool{\mathcal{P}^{L}}
\def\LastLabelPool{\mathcal{P}^{L}(r-1)}
\def\CurLabelPool{\mathcal{P}^{L}(r)}

\def\CurPreProcessPool{\mathcal{P}^{R}(r)}

\def\LastUnlabelPoolCLS{\mathcal{P}_{C}^{U}(r-1)}

\def\LastUnlabelPoolDET{\mathcal{P}_{D}^{U}(r-1)}

\def\LastLabelPoolCLS{\mathcal{P}_{C}^{L}(r-1)}

\def\CurPreProcessPoolCLS{\mathcal{P}_{C}^{R}(r)}

\def\CurPreProcessPoolDET{\mathcal{P}_{D}^{R}(r)}

\def\CurLabelSet{\mathcal{S}^{L}(r)}

\def\UnlabelSet{\mathcal{S}^{U}}

\def\CurUnlabelSet{\mathcal{S}^{U}(r)}

\def\LastModel{\phi(r-1)}

\def\LastModelCLS{\phi_{C}(r-1)}

\def\LastModelDET{\phi_{D}(r-1)}

\def\WikiLabelNum{170,586}

\def\AllLabelNum{304,048}

\def\CommonLabelNum{809}

\def\DetTotalNum{28M}
\def\ClsTotalNum{69M}

\begin{document}
\title{Bamboo: Building Mega-Scale Vision Dataset Continually with Human-Machine Synergy}
%
%
%
%

\author{Yuanhan Zhang, Qinghong Sun, Yichun Zhou, Zexin He, \protect\\ Zhenfei Yin, Kun Wang, Lu Sheng, Yu Qiao, Jing Shao, Ziwei Liu
\IEEEcompsocitemizethanks{\IEEEcompsocthanksitem Yuanhan Zhang and Ziwei Liu are with the S-Lab, Nanyang Technological University \protect \\
E-mail: \{yuanhan002, ziwei.liu\}@ntu.edu.sg
\IEEEcompsocthanksitem Qinghong Sun, Zhenfei Yin, Kun Wang and Jing Shao are with SenseTime Research.
E-mail: \{sunqinghong, yinzhenfei, wangkun, shaojing\}@senseauto.com
\IEEEcompsocthanksitem Yichun Zhou, Zexin He, Lu Sheng are with Beihang University.
E-mail: \{buaazyc, jacquesdeh, lsheng\}@buaa.edu.cn 
\IEEEcompsocthanksitem Yu Qiao is with Shanghai AI Laboratory.
E-mail: qiaoyu@pjlab.org.cn}}

%
%

\markboth{Journal of \LaTeX\ Class Files,~Vol.~14, No.~8, August~2015}%
{Shell \MakeLowercase{\textit{et al.}}: Bare Demo of IEEEtran.cls for Computer Society Journals}
%



\IEEEtitleabstractindextext{%
\begin{abstract}
Large-scale datasets play a vital role in computer vision. But current datasets are annotated blindly without differentiation to samples, making the data collection inefficient and unscalable. The open question is how to build a mega-scale dataset actively. Although advanced active learning algorithms might be the answer, we experimentally found that they are lame in the realistic annotation scenario where out-of-distribution data is extensive. This work thus proposes a novel active learning framework for realistic dataset annotation. Equipped with this framework, we build a high-quality vision dataset---\textbf{\Datasetname}, which consists of 69M image classification annotations with 119K categories and 28M object bounding box annotations with 809 categories. We organize these categories by a hierarchical taxonomy integrated from several knowledge bases. The classification annotations are four times larger than ImageNet22K, and that of detection is three times larger than Object365. Compared to ImageNet22K and Objects365, models pre-trained on \Datasetname~achieve superior performance among various downstream tasks (6.2\% gains on classification and 2.1\% gains on detection). We believe our active learning framework and Bamboo are essential for future work. Code and dataset are available at \url{https://github.com/ZhangYuanhan-AI/Bamboo}.
\end{abstract}

}

\maketitle

\IEEEdisplaynontitleabstractindextext

%
\IEEEpeerreviewmaketitle

\input{sections/2.introduction.tex}
\input{sections/3.RelatedWork.tex}
\input{sections/4.Taxonomy.tex}
\input{sections/5.Pipeline.tex}
\input{sections/6.DatasetStatistics.tex}
\input{sections/7.AlgorithmicAnalysis.tex}

\input{sections/10.social_impact}

\input{sections/8.Discussion.tex}
{
\bibliographystyle{plain}
\bibliography{sections/submission}
}

%








\end{document}

%% file: sections/2.Introduction.tex
\section{Introduction}
\label{Introduction}


\begin{figure*}[t]
    \hsize=\textwidth
    \centering
    \includegraphics[width=\linewidth]{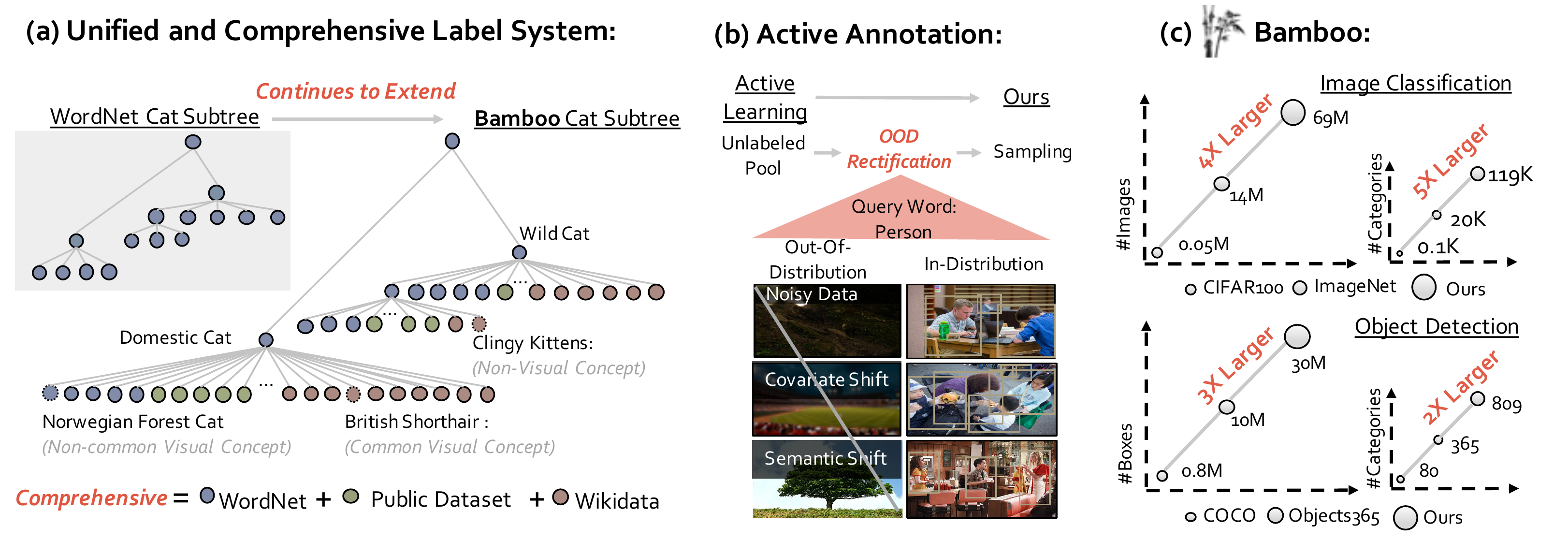}
    \caption{\textbf{The overview of \textit{Bamboo} Dataset.} Bamboo is a new mega-scale vision dataset built on a comprehensive label system with human-machine synergy. (a) Our label system continually extends from WordNet with our solutions. Concepts in the label system are distinguished as ``common visual'', ``non-common visual'' or ``non-visual'' concepts. (b) Raw data crawled by the query word \textit{person} includes both the in-distribution (ID) data and out-of-distribution (OOD) data. OOD data implies noisy, covariate shift, and semantic shift data. Noisy data does not present useful semantic information. Covariate shift data implies semantic information, \ie person. However, such semantic information is of poor quality, annotators thus hard to annotate. Semantic shift data also implies sxemantic information, \ie tree. But the tree is not related to the query word person. \textit{OOD rectification} mitigates the ineffectiveness of active learning through filtering OOD data. (c) Bamboo collects \ClsTotalNum~classification annotation and \DetTotalNum~bounding box annotations.}
    \label{fig:fig1}
\end{figure*}

\IEEEPARstart{L}{arge}-scale pre-trained models, either trained in a supervised~\cite{multi-task,bit,SWSL} or unsupervised~\cite{MOCOv1,MOCOv2,swav} manner, have become a foundation for modern computer vision.
Pre-trained models~\cite{CLIP} bring various applications by transferring to downstream tasks.
Most importantly, the fuel of this foundation models~\cite {foundationmodel} relies on the availability of increasingly large and diverse datasets~\cite{cifar,deng2009imagenet,JFT300M,COCO}. 

Building a high-quality dataset requires careful consideration in selecting data.
However, public datasets are annotated blindly with no differentiation to samples, which brings a colossal waste of annotation budget: Citovsky \etal~\cite{citovsky2021batch} indicates that only 70\% data in OpenImages~\cite{kuznetsova2020open} can achieve on-par performance to its complete set. Though active learning (AL) researchers extensively study how to select the most valuable samples---informative and in-distribution--from unlabeled data pool~\cite{AL1,AL3,settles2009active,AL5,AL6,AL7,AL8,AL10,gal2016uncertainty},  we experimentally observe that the current advanced active learning methods, \eg ClusterMargin~\cite{citovsky2021batch}, Margin~\cite{Margin-Based} and CoreSet~\cite{coresetorigin},  are lame in the realistic annotation scenario. 
Specifically, AL methods select high-informative data at the expense of choosing out-of-distribution data discarded by annotators and not used for model supervised learning.
Random sampling selects less high-informative data than AL but includes much more in-distribution data than AL. As the performance gain bought by data quantity is superior to data quality when annotated data of AL is 70\% less than that of random sampling (as shown in Fig.~\ref{fig:Bamboo-AL-CLS}), AL is inferior to random sampling for improving model performance.
Given this shortage, we propose a novel active learning framework, which cleans off the out-of-distribution data in the unlabeled data pool before active sampling, ensuring the sampled data under active learning is informative and meanwhile in-distribution. This novel framework achieves better than random sampling for boosting supervised learning model performance. 

We aim to annotate a mega-scale classification and object detection dataset with our proposed active learning framework. First, we build a comprehensive label system for querying diverse data covering numerous semantics. Specifically, we form a unified label system with a hierarchical structure consisting of \AllLabelNum~categories.
It integrates label systems from 19 latest public image classification datasets and five object detection datasets and also absorbs \WikiLabelNum~new categories from knowledge bases, \eg Wikidata~\cite{wikidata}.
Then, we contribute \textbf{\Datasetname~Dataset}, a mega-scale and information-dense dataset for the pre-training of both classification and detection, which is active annotating by human-machine synergy. \Datasetname~has three appealing properties:

\begin{itemize}[leftmargin=*]

\item \textbf{Comprehensive.} 
It consists of \ClsTotalNum~image classification annotations
and \DetTotalNum~object bounding box annotations, spanning over 119K visual categories.
The scale of the label system and the annotated data are the largest among all the publicly available datasets. We illustrate the comparison of Bamboo and publicly available datasets in the Fig.~\ref{fig:fig1}(c).
%

\item \textbf{Information-Dense.} 
We guarantee Bamboo is highly informative through the label system and the annotated data. The label system is constructed by thoroughly integrating public datasets and knowledge bases. 
Our active annotation pipeline specifically selects the annotated data to reduce model uncertainty.

\item \textbf{Continual.}
Our label system keeps the dataset growing with the automatic concept linking strategies. We can constantly absorb new categories in the real world and integrate them into our label system. Moreover, leveraging the ever-increasing internet data, our active annotation pipeline will steadily sustainably expand the \Datasetname~dataset size.
\end{itemize}

Extensive experiments demonstrate that \Datasetname~dataset is an effective pre-training source.
The \Datasetname~pre-trained model significantly outperforms CLIP ViT B/16~\cite{CLIP} pre-trained model with 6.2\% gain (85.6\% vs 91.8\%) on classification, and outperforms Objects365~\cite{shao2019objects365} pre-trained model with 2.1\% gain (14.7\% vs 12.6\%) on CityPersons~\cite{CityPersons}.
In addition, we provide valuable observations regarding large-scale pre-training from over 1,000 experiments.
We hope the \Datasetname~dataset and these observations will pave the way for developing more general and effective vision models.

%% file: sections/3.RelatedWork.tex
\section{Related Works}
\label{Related Works}

\noindent \textbf{Learning of Representation at Scale.}
Representation learning has advanced thanks to improvements in various learning paradigms and large-scale datasets. 
Supervised learning models~\cite{resnet, Xception} leverage label information to supervise representation learning, achieving excellent performance among various downstream tasks.
To avoid the need for annotations that require a tremendous amount of human and labeling cost, weakly-supervised and self-supervised pre-training methods have been proposed.
Self-supervised methods~\cite{deepcluster,PIRL,MOCOv1,simclr,swav, Colorization, Colorization_2, instance_1, instance_2, instance_3} with contrastive learning have shown that unsupervised pre-training produces features surpassing the supervised feature representations on many downstream tasks~\cite{cifar,flowers,pets,sun,stanfordcar}.
Large weakly-supervised datasets, such as Instagram hashtags~\cite{IN-1B} and JFT~\cite{JFT300M}, helps model~\cite{SWSL, self-training,self-training_2} achieve significant gains on downstream tasks. 
In addition, CLIP~\cite{CLIP} pre-train models on both the image signal and text signal, achieving good performance for the zero-shot evaluation.
Our study is part of a larger body of work on training models on sizeable supervised image datasets. As the labeling cost that hurdles the supervised learning dataset is becoming increasingly significant, we systematically investigate how to collect, annotate and build a mega-scale dataset efficiently, actively and continually.

\noindent \textbf{Active Learning.}
Active learning (AL) aims at finding the minimum number
of labeled images to have a supervised learning algorithm reach a certain performance~\cite{AL1,AL3,settles2009active,AL5,AL6,AL7,AL8,AL10,gal2016uncertainty}. The main component in an active learning loop is sampling strategies. 
%
%
The existing AL research is conducted on the curated datasets. Each data point in the labeled and unlabeled pool of these datasets is valid, \ie available for labeling. 
However, curated datasets can hardly imitate the annotation in realistic scenarios where out-of-distribution data that is unavailable for labeling appears on a large scale in the unlabeled pool. 
From our experiments, we find the existing AL methods lag in realistic scenarios. Therefore, we propose a novel active annotation pipeline to improve the performance of AL methods in realistic scenarios.

%% file: sections/4.Taxonomy.tex
\section{Label System Construction}
\label{GVB}

In this section, we briefly introduce how to build a comprehensive label system. The number of concepts decides the data amount upper bound---we crawl data based on querying these labels. Starting from WordNet~\cite{wordnet}, we enrich its concepts from another two concept resources (Sec.~\ref{Integrating Existing Concept}) through three designed linking strategies (Sec.~\ref{Concepts Integration})).

\subsection{Concepts Resources}
\label{Integrating Existing Concept}
\noindent \textbf{WordNet.} WordNet is a lexical database of semantic relations between concepts in more than 200 languages. Each meaningful concept in WordNet, possibly described by multiple words or phrases, is called a ``synset''. Referring to ImageNet22K~\cite{deng2009imagenet}, we only use the Noun words of WordNet. WordNet is the foundation of our label system.

\noindent \textbf{Public Datasets.} We collect 24 public datasets, including ImageNet22K~\cite{deng2009imagenet}, OpenImages~\cite{kuznetsova2020open}, COCO~\cite{COCO}, iNaturalist~\cite{inaturalist}, and \emph{etc}.\footnote{The complete list of public datasets is reported in \textit{Supplementary Material}.} These datasets cover a wide range of datasets in both image classification and object detection.

\noindent \textbf{Wikidata.} Wikidata~\cite{wikidata} contains a large number of concepts, such as different kinds of foods, animals, and locations. As the number of concepts in Wikidata continues to grow, so far, we have included \WikiLabelNum~concepts from it. These concepts are the leaf nodes in their taxonomy.

\subsection{Concepts Integration}
\label{Concepts Integration}
WordNet is a lexical graph whose concepts imply semantic relation. For example, the father node of ``British Shorthair'' is ``Domestic Cat''. How to integrate concepts from public datasets and Wikidata into this WordNet is an open question.
We propose three parallel solutions to integrate these categories into WordNet in this work. 
%


\noindent\textbf{Solution 1: Leveraging on the \emph{subclassOf}.} 
The taxonomy of Wikidata is contributed by adding the ``subclassOf'' that is related to the hypernyms relationship in the taxonomy of WordNet. Referred to~\cite{Tanon2020YAGO4A}, we link Wikidata leaf node concepts to the WordNet by leveraging the ``subclassOf''.

\noindent\textbf{Solution 2: Parsing the Concept.} 
Referred to the previous work~\cite{fabian2007yago}, we can also link the concept to the WordNet~through word parsing. 
For example, for the concept \textit{Sumatran Orangutan}, we parse this concept~\cite{spacy2} and get its head compound ``Orangutan''. 
%
In this way, we add \textit{Sumatran Orangutan} as the new hyponym of the ``Orangutan'' if ``Orangutan'' exists in WordNet.

\noindent\textbf{Solution 3: Linking to the Closed Synset.}  We calculate the word embedding of both the synsets and given concepts through Spacy~\cite{spacy2}. 
If a given concept cannot be linked to WordNet, we add this category to the hyponym of its nearest cosine distance synset.

\begin{figure}[t]
\centering
\includegraphics[width=0.45\textwidth]{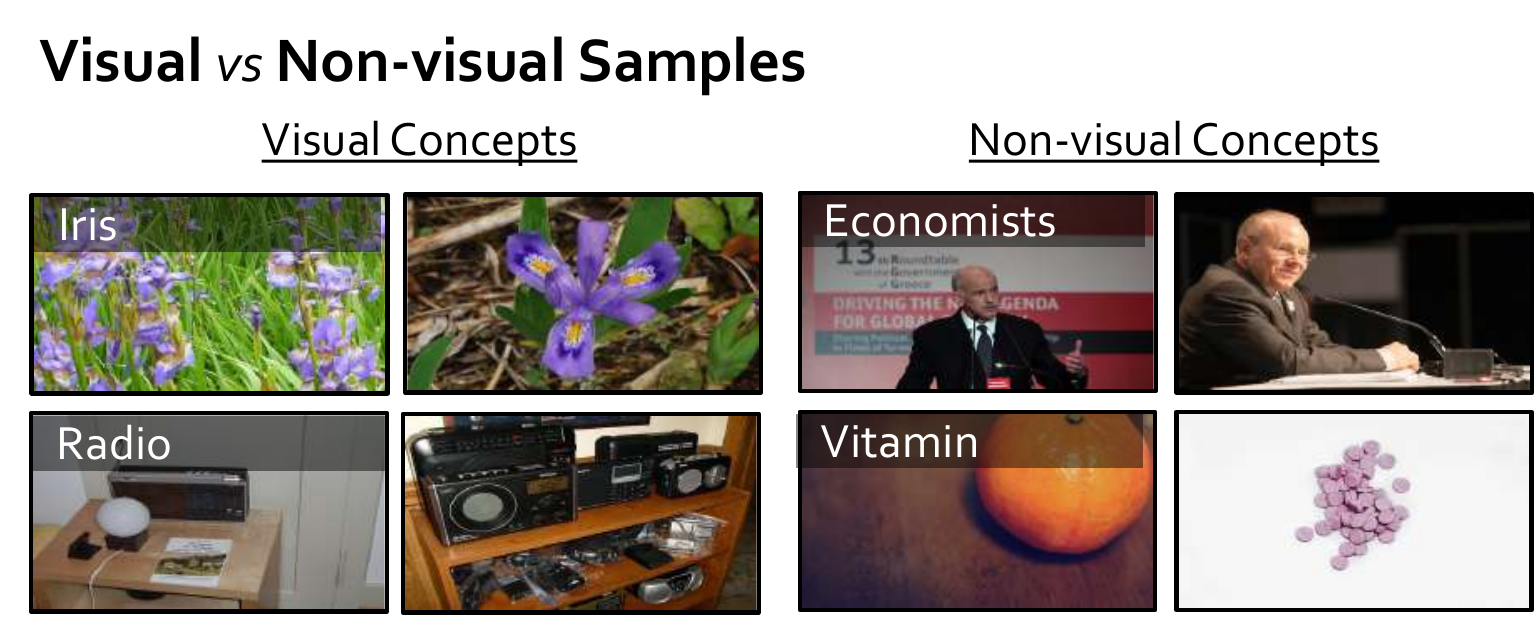}
\caption{\textbf{The illustration of visual and Non-visual concept.} \textit{Vitamin} do not share any common semantic information.  \textit{Economists} implies common semantic information---Man---but 
economists are not necessarily men.
}
\label{fig:visual_and_nonvisual}
\end{figure}

\begin{figure*}[t]
\centering
\includegraphics[width=\textwidth]{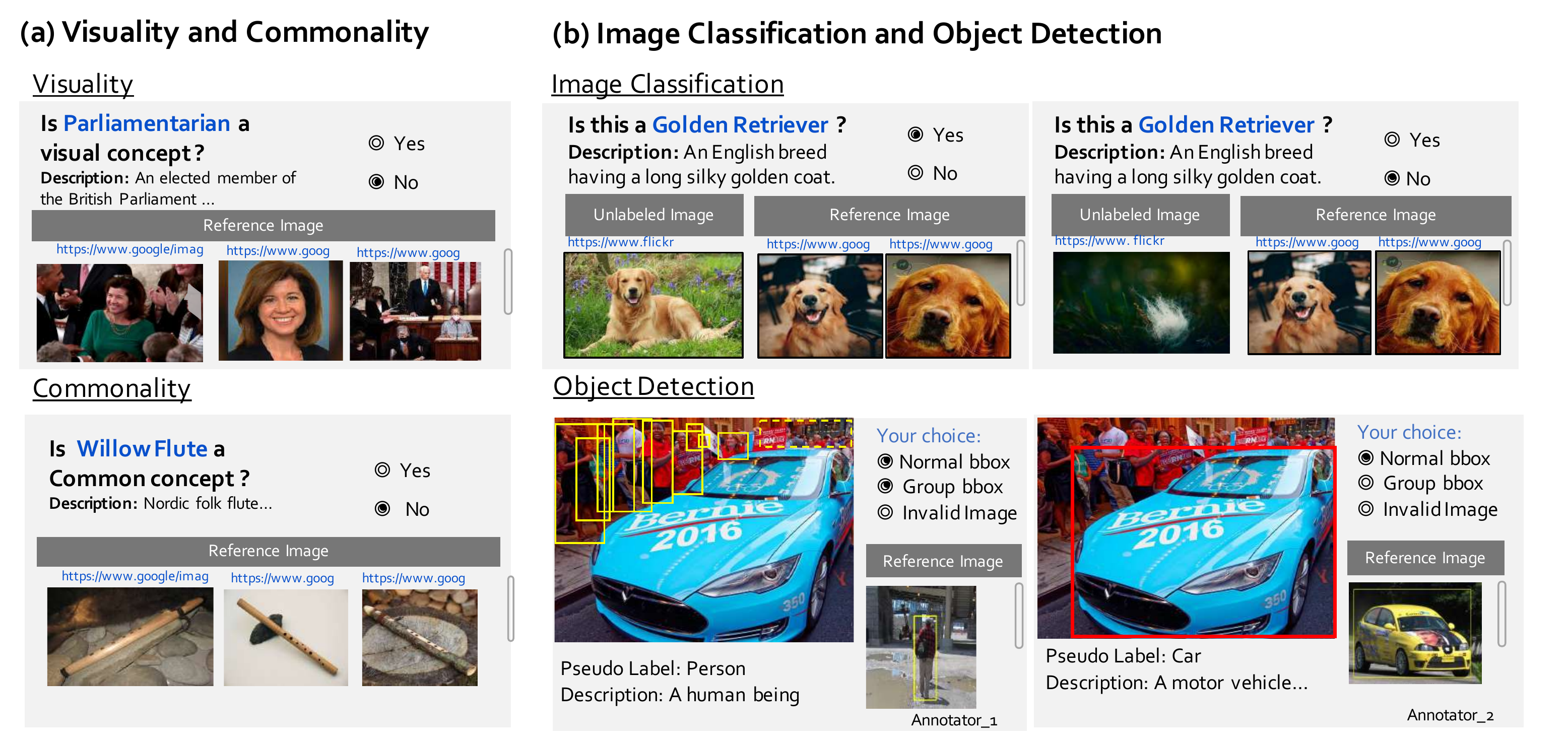}
\caption{\textbf{User interfaces for concept tagging and annotation.} (a) The meta information of the concept tagging consists of tags, descriptions, and reference images.
(b) Interface for image classification and object detection. For the object detection task. The image is assigned to different annotators based on its multiple pseudo labels. In addition, annotators should choose the attribute of the bounding box. The criteria for the attribute options are described in detail in \textit{Supplementary Material}.}
\label{fig:concept_tagging_main}
\end{figure*}

\subsection{Concepts Tagging}
\label{Concepts Tagging}

\noindent\textbf{Visuality.}
Yang \etal has mentioned the non-visual category problem in their work~\cite{yang2020towardsfairer}. We illustrate visual and non-visual words in Fig.~\ref{fig:visual_and_nonvisual}.
To mitigate this problem, we conduct visual concept tagging for our build label system. Specifically, a concept is non-visual if three out of five annotators think this word is less concrete, and its sample images can rarely imply a common semantic meaning. We illustrate the concept tagging in Fig.~\ref{fig:concept_tagging_main}(a).

\noindent\textbf{Commonality.}
Based on the visual concepts, we further conduct common concept annotation for all visual concepts. Referred to COCO~\cite{COCO}, “common concept” refers to entry-level categories that are commonly used by
humans when describing objects (\eg dog, chair, person). Specifically, a concept is positive only if it receives at least three-fifths of the votes. Based on the proposed annotation method, we retain \CommonLabelNum~common concepts for the annotation of object detection.

%% file: sections/5.Pipeline.tex
\section{Active Dataset Construction - \Datasetname}
\label{Active Annotation Pipeline}

\begin{figure*}[t]
\centering
\includegraphics[width=\textwidth]{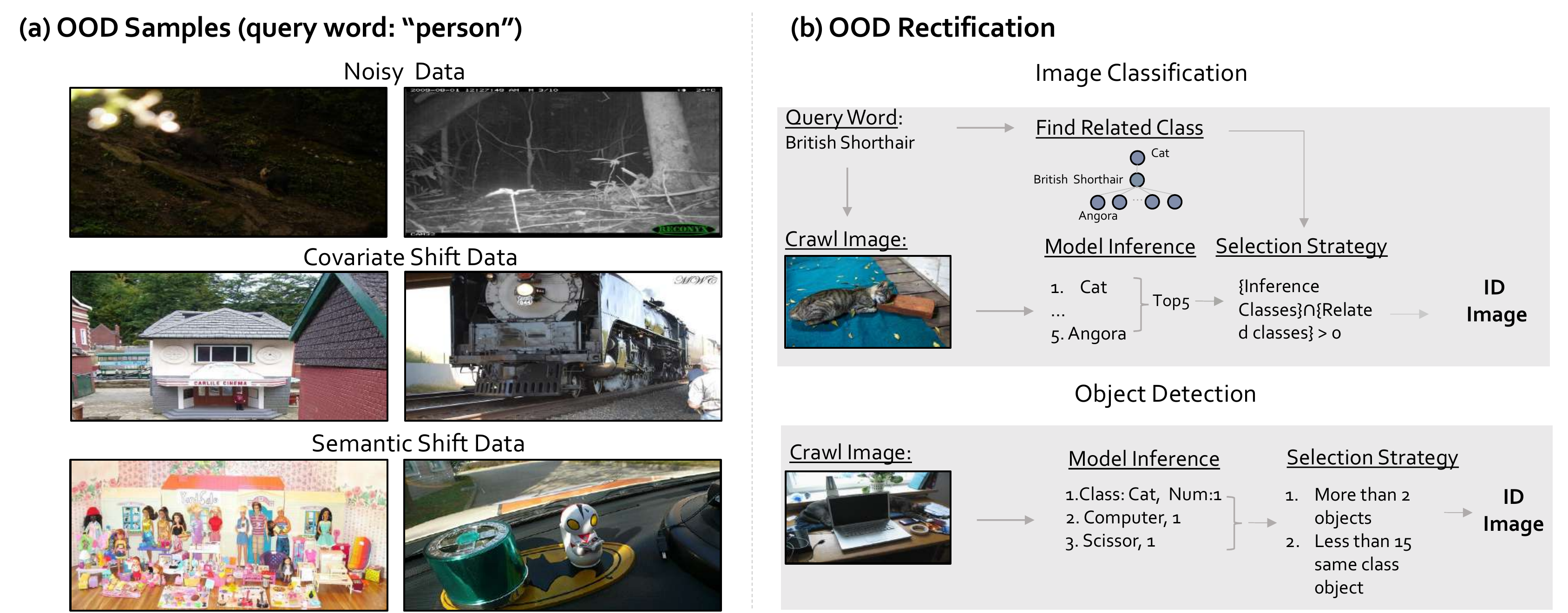}
\caption{\textbf{(a) The illustration of out-of-distribution (OOD) data in realistic scenarios.} Mainly, three types of OOD data exist in the unlabeled data pool, including noisy data, covariate shift data (\ie, OOD samples from a different domain), and semantic shift data (\ie, OOD samples are drawn from different classes).  \textbf{(b) The illustration of OOD rectification.} OOD rectification filters OOD data in the unlabeled data pool, which is crucial for active learning.
}
\label{fig:ood}
\end{figure*}

Equipped with the unified and comprehensive label system, we start to construct \Datasetname actively. In this section. We first introduce the active learning pipeline for building Bamboo in Sec~\ref{Active Learning Framework}. We summarize this pipeline in Algorithm~\ref{Alg:Dataset Construction Pipeline}.
Then in Sec.~\ref{sec:Simulated experiments For Active Learning}, we discuss the superiority of our newly proposed active learning methods---we are the \textbf{first} time beat the random sampling in selecting the most valuable data for data pre-training.

\IncMargin{1em}
\begin{algorithm}[h]
\caption{Outline of AL Framework}
\label{Alg:Dataset Construction Pipeline}
  \SetKwInOut{Input}{input}\SetKwInOut{Output}{output}

\Input{Raw unlabeled pool $\varTheta$;
Number of active learning rounds \textit{T};
Neural network $\phi$;
}


$\LabelPool(0) \leftarrow$ Annotating a few data from $\varTheta$ and adding all inherited data as cold start;

$\UnlabelPool(0)$ $\leftarrow$ $\varTheta - \LabelPool(0) \cap \varTheta$\;

Initializing model $\phi(0)$ based on $\LabelPool(0)$\;
\BlankLine
  \For{$r\leftarrow 1$ \KwTo $T$}{
    \footnotemark $\CurPreProcessPool$ $\leftarrow$ Rectifying $\LastUnlabelPool$ w/ $\LastModel$\;
    $\CurUnlabelSet$ $\leftarrow$ Sampling in $\CurPreProcessPool$ w/ $\LastModel$\;
    $\CurLabelSet$ $\leftarrow$ Annotating valid data from $\CurUnlabelSet$\;
    $\CurUnlabelPool$ $\leftarrow$ $\LastUnlabelPool - \CurUnlabelSet$\;
    $\CurLabelPool$ $\leftarrow$ $\LastLabelPool \cup \CurLabelSet$\;
    Training $\phi(r)$ on $\CurLabelPool$\;
  }
\end{algorithm}
\DecMargin{1em}
\footnotetext{This step is not included in the current active learning
research.}

\subsection{Active Learning Framework}
\label{Active Learning Framework}
\subsubsection{Building Unlabeled Data Pool}
\label{Data Collection}
For image classification, one query word has one visual concept mentioned in Sec.~\ref{Concepts Tagging}. For object detection, one query has two concepts, \ie, common concept +  scene semantic word or common concept + common concept. For example,  \textit{dog} + \textit{street} or \textit{dog} + \textit{ball}. To further enrich the searching results, any given query word can be converted to its synonyms or its Chinese, Spanish, Dutch and Italian version for querying.
Totally, we obtain a 170M unlabeled pool for classification and a 200M unlabeled pool for detection.

\subsubsection{Cold Start }
\label{sec:Inheriting Public Datasets}
Cold start is the very first step for active learning. The labeled date pool $\LabelPool(0)$ to initialize the model \textbf{$\phi(0)$} for the cold start phase include two parts as follows.

\noindent \textbf{Public Dataset.} As mentioned in Sec.~\ref{Integrating Existing Concept}, we use 24 datasets as concept resources, including 19 image classification datasets and 5 object detection datasets.
Refereed to the evaluation suite of popular transfer learning study~\cite{ImageNettransferbetter_1,VTAB,MOCOv1}, we select 12 datasets for downstream evaluation. We include the annotation of the other 12 datasets---9 image classification datasets and 3 object detection datasets. 
In total, we inherit 27,848,477 classification annotations and 21,983,223 object bounding box annotations from those 12 datasets.

\noindent \textbf{New Annotated Data.} For concepts not included in public datasets, we sample images from unlabeled pool $\varTheta$ and annotate data for them until they have 50 annotated data.

\subsubsection{OOD Rectification} 
\noindent \textbf{Image Classification.}
In this step, we rectify the latest unlabeled data pool $\LastUnlabelPoolCLS$.
As shown in Fig~\ref{fig:ood} (b), in each round $r$, we firstly utilize $\LastModelCLS$ trained on $\LastLabelPoolCLS$~to acquire predictions of each image in $\LastUnlabelPoolCLS$. 
We infer an image is out-of-distribution if its top-5 predicted categories do not overlap with its related categories.
Specifically, we define the related categories of an image as the sub-population of hypernyms of its query word. 
If an image is not out-of-distribution, we add it into $\CurPreProcessPoolCLS$~for further data sampling.
In Sec.~\ref{sec:Simulated experiments For Active Learning}, we empirically observe that OOD rectification is essential for AL to function in realistic scenarios.

\noindent \textbf{Object Detection.}
Similar to classification, we acquire proposal predictions of each image in $\LastUnlabelPoolDET$ by $\LastModelDET$. 
On the one hand, we filter out the image with less than two proposals. Such images might be noisy data or semantic shift data.
On the other hand, we filter out the image with more than 15 identical semantic proposals since such image might be the covariate shift data. 
As shown in Fig~\ref{fig:ood} (b), the remaining in-distribution data forms $\CurPreProcessPoolDET$ for the data sampling.

\subsubsection{Data Sampling.}
In this step, we use ClusterMargin~\cite{citovsky2021batch}, which considers both the uncertainty and diversity in data, to select the most valuable data $\CurUnlabelSet$ from the latest rectified data pool $\CurPreProcessPool$ for annotation. 

\subsubsection{Data Annotation.}
\label{sec:Data Annotation}
Finally, we rely on an online platform to annotate valid data---its querying word accurately describes the semantic meaning of this data---in $\UnlabelSet(r)$, forming the labeled data set $\CurLabelSet$. We illustrate our online platform in Fig.~\ref{fig:concept_tagging_main}, and introduce the details of annotations as follows. 
\noindent \textbf{Image Classification.}
To provide a comprehensive definition of each category, we construct reference images that are collected by querying Google image search and Wikipedia~\cite{wiki:xxx}. 
For each image in $\CurUnlabelSet$, its meta-information has two parts: the query word of this image and the reference images of the query word. 
We then ask the five annotators whether this image conforms to its meta information. An image is annotated and added into $\CurLabelSet$---valid data---only if at least 3 out of 5 annotators give the positive answer to the question as mentioned above.

\noindent \textbf{Object Detection.}
Following Objects365~\cite{shao2019objects365},  
one annotator is responsible for annotating a specific category, which improves the annotation efficiency and quality.
Similar to OpenImages~\cite{kuznetsova2020open}, meta information of an image includes not only its reference images but also its pseudo labels that include i) the query words of this image. ii) the category predictions of available detection models. iii) re-labeling predictions~\cite{Re-labeling} of the latest trained classification model $\phi_{C}$.

\begin{figure}[t]
    \centering
    \includegraphics[width=0.95\linewidth]{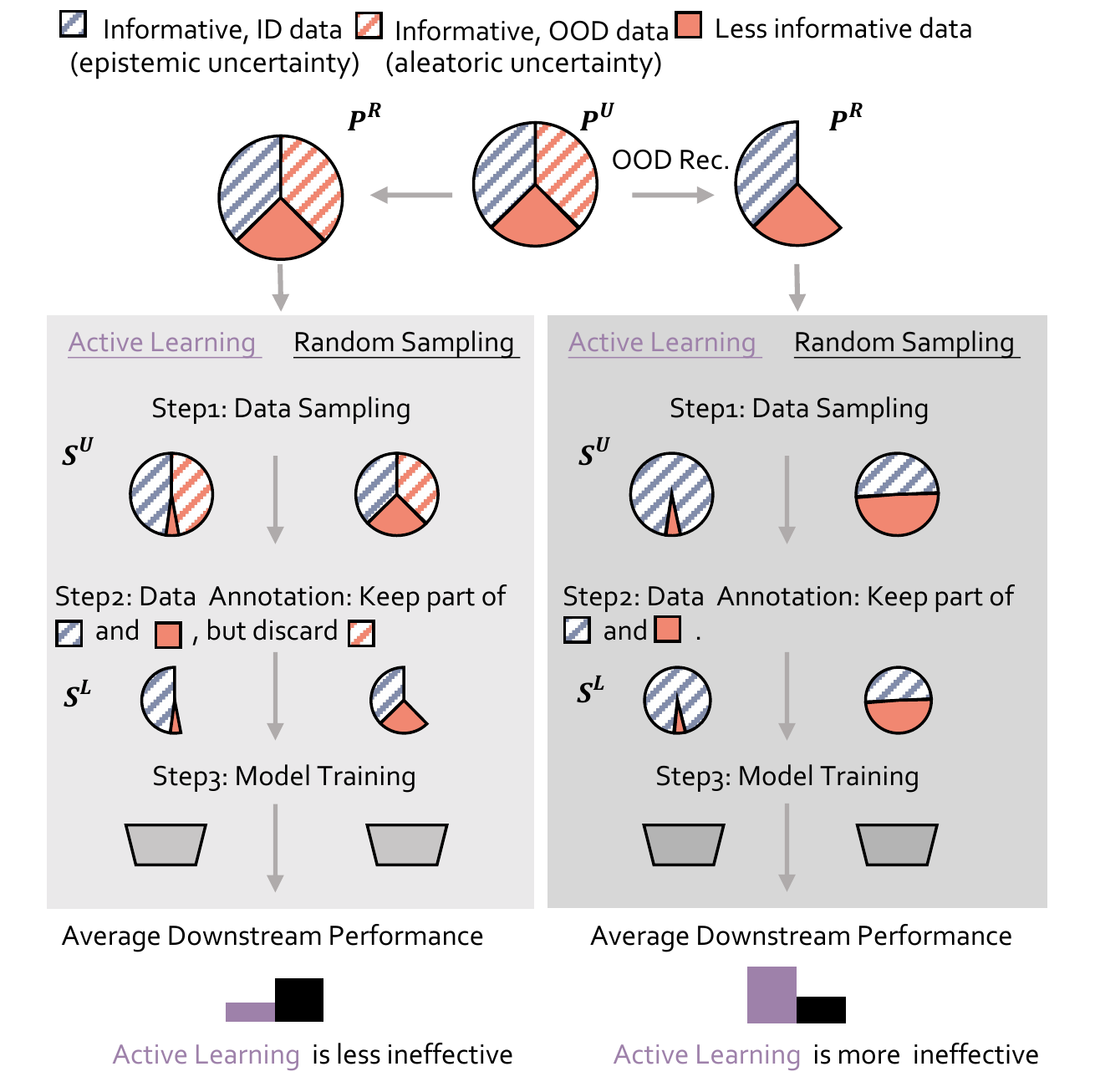}
    \caption{The Illustration of how our OOD rectification step helps active learning performs better in realistic scenarios. }
    \label{fig:Why_not_work}
\end{figure}

\begin{figure*}[t]
\centering
\includegraphics[width=\textwidth]{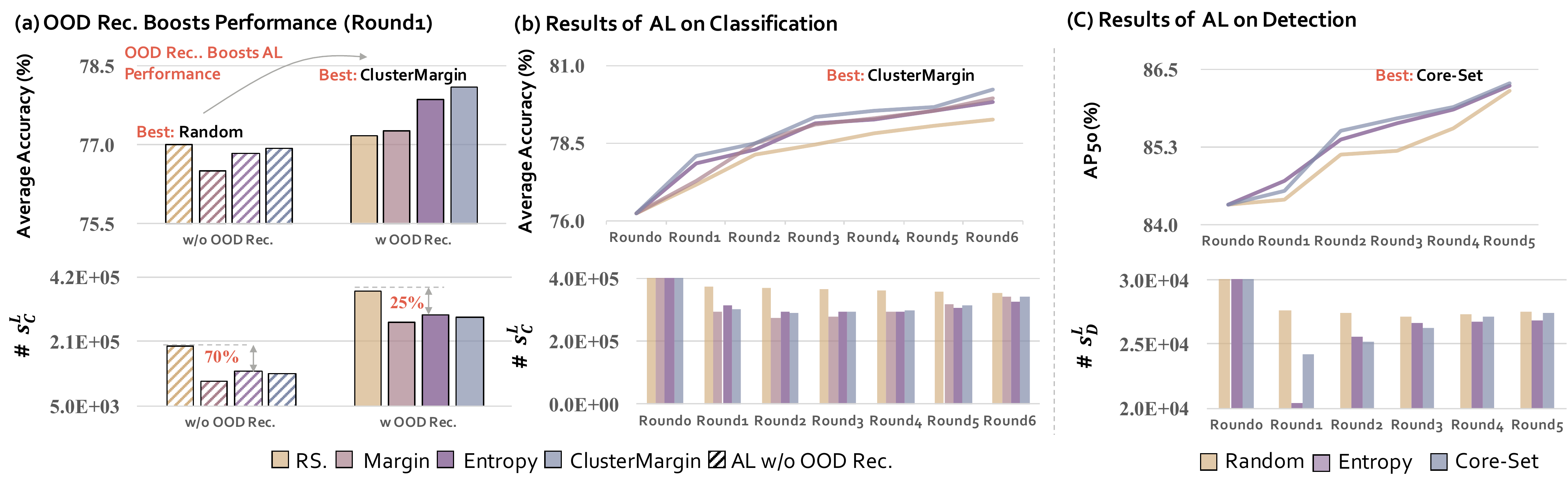}
\vspace{-20pt}
\caption{\textbf{The study of active annotation in Bamboo}. \textbf{(a)} current AL methods struggle in realistic scenarios. Random sampling achieves better performance than each AL method. \textit{OOD Rectification} boosts all AL methods to outperform random sampling. AL methods are still more helpful for model training with less valid data. It implies that the valid data that AL methods selected are much more informative. \textbf{(b) and (c)} in both classification and detection tasks, AL methods (ClusterMargin and Core-Set) that consider both the uncertainty and diversity select the most valuable data for model training. $S_\textit{C}^\textit{L}$ refers annotated valid data from a given AL batch. Average accuracy denotes the average performance of models on the downstream datasets.
}
\label{fig:Bamboo-AL-CLS}
\vspace{-15pt}
\end{figure*}

\subsection{Studies on Active Annotation}
\label{sec:Simulated experiments For Active Learning}
In academic active learning (AL) works~\cite{citovsky2021batch,semiactivelearning}, researchers conduct data sampling on the leave-out ``unlabeled'' data pool that are separated from a curated dataset, \eg ImageNet~\cite{citovsky2021batch} and CIFAR10~\cite{Margin-Based}.  All the data in this ``unlabeled'' data pool is strictly valid.\footnote{Annotator had deleted invalid data as dataset established.} However, in realistic annotation scenarios, the real unlabeled data pool is composed of valid data and invalid data that is mostly out-of-distribution data, as shown in Fig.~\ref{fig:ood}(a). Therefore, can AL methods are effective when the invalid data is in the unlabeled data pool is an open question.
And we found that:

\textit{Current Active Learning Methods are Ineffective for Sampling Valuable Data in the Real unlabeled data pool.}

As shown in Fig.~\ref{fig:Bamboo-AL-CLS}(a), we illustrate the number of $\mathcal{S}^{L}(1)$  (the first round valid data set of Bamboo) when $\mathcal{P}^{R}(1) \leftarrow \mathcal{P}^{U}(0)$ (the academic active learning framework). We observe that AL sampling would retain fewer data in $\mathcal{S}^{L}(1)$ than random sampling.
For example, Entropy Sampling selects 70\% less data than random sampling, resulting in worse downstream performance.

\noindent \textbf{The Devils are in Uncertainty Modeling.} 
As discussed in~\cite{kendall2017uncertainties, d2021tale}, there are mainly two types of uncertainty for the deep models: \textit{Aleatoric} and \textit{Epistemic}.
Both uncertainties are informative, but the aleatoric uncertainty is the out-of-distribution data, and the epistemic uncertainty is the in-distribution data.
Considering $\mathcal{P}^{U}(0)$ where aleatoric-uncertain data, epistemic-uncertain data, and other less-informative data exist,  
when $\mathcal{P}^{R}(1) \leftarrow \mathcal{P}^{U}(0)$, $\mathcal{S}^{U}(1)$ under AL sampling would have more aleatoric-uncertain data than that under random sampling, as AL methods tend to select uncertain data. 
Eventually, $\mathcal{S}^{L}(1)$ under AL sampling should has less data than that under random sampling as aleatoric uncertain data is invalid for annotators. We illustrate this phenomenon in Fig.~\ref{fig:Why_not_work} left. As shown in Fig.~\ref{fig:Bamboo-AL-CLS} (a), with much less $\mathcal{S}^{L}(1)$, AL methods' performances are hence worse than RS. 

\noindent \textbf{OOD Rec. Boosts AL Performance.} 
When $\mathcal{P}^{R}(1) \leftarrow \text{Rectifying}\ \mathcal{P}^{U}(0)\ \text{w/ }\phi(0)$ (our active learning framework), our proposed OOD rectification filters out the aleatoric uncertain data in $\mathcal{P}^{U}(0)$. 
Therefore, $\mathcal{P}^{R}(1)$ is only comprised of epistemic-uncertain data---which is informative---and other less-informative data. Since AL methods would select more epistemic uncertain data in $\mathcal{P}^{R}(1)$ than random sampling, they eventually perform better. We illustrate how OOD rectification helps active
learning performs better in realistic scenarios in Fig.~\ref{fig:Why_not_work} right.
As shown in Fig.~\ref{fig:Bamboo-AL-CLS}(b,c), with OOD rectification, in both classification and detection tasks, AL methods (ClusterMargin and Core-Set) that consider both the uncertainty and diversity select the most valuable data for model training.

%% file: sections/6.DatasetStatistics.tex
\section{Dataset Statistics}
\label{Dataset Statistics}

\begin{figure*}[t]
\centering
\includegraphics[width=\textwidth]{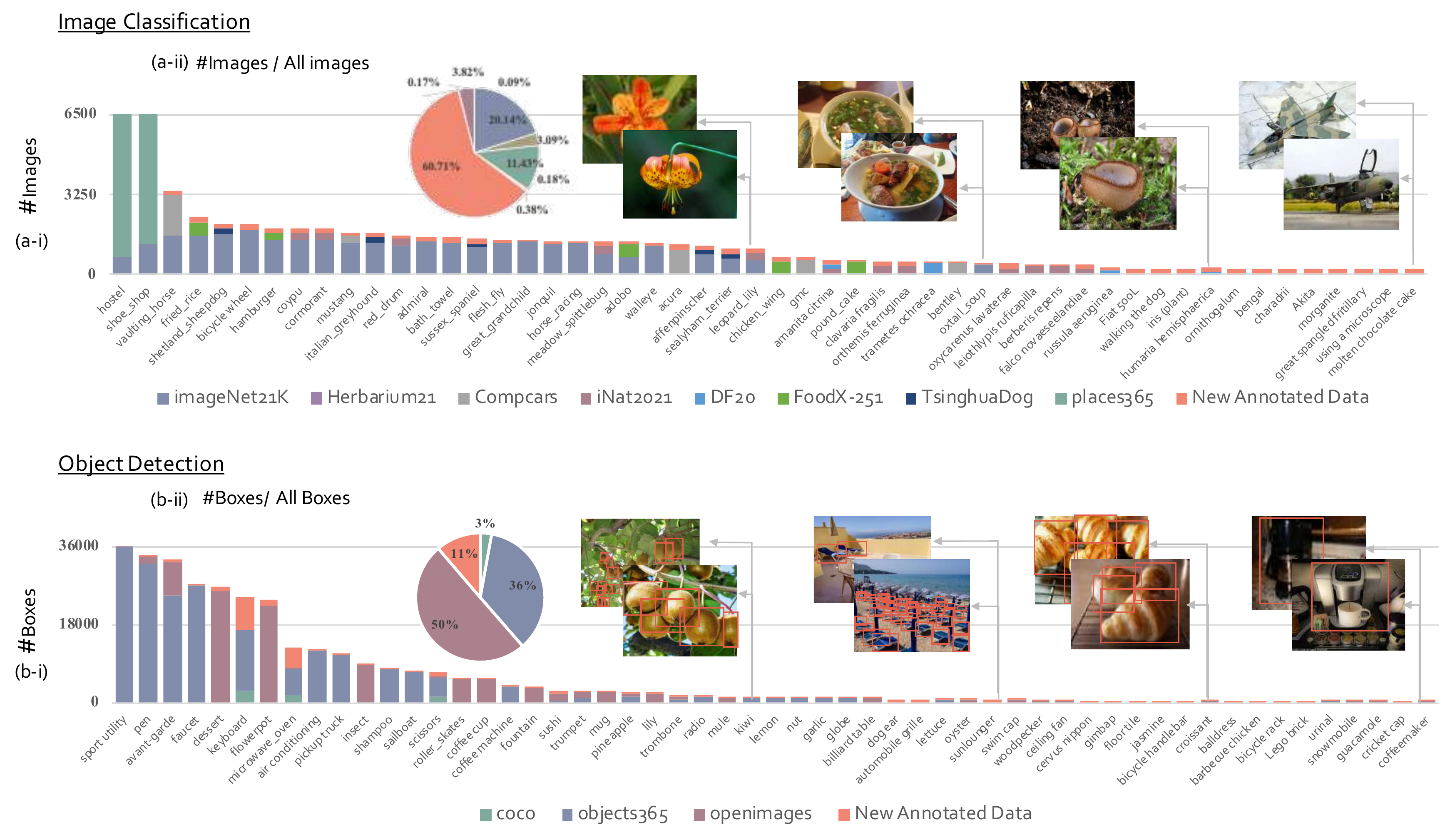}
\caption{\textbf{Sorted distribution of image number per category in the \Datasetname}. (a-i) \Datasetname-CLS~contains 68,884,828 images spread across 119,035 categories. Category names are shown for every 250 intervals. \Datasetname-CLS includes some fine-grained concepts that not be included in the current public datasets, such as \textit{Folland Midge}. (a-ii) The new classification annotated data accounts for 60.71\% of images in \Datasetname.
(b-i) \Datasetname-DET~contains 3,104,012 images across 809 categories. Category names are shown for every 16 intervals. (b-ii) The new detection annotated data accounts for 11\% of images in \Datasetname.
}
\label{fig:statistics}
\end{figure*}

\begin{table*}[t]
\small
\caption{\textbf{Left: The statistics of the number of
bounding boxes per image.} Quantitatively, our new annotated data has 8.3 instances (on average) per image, which is more dense compared with the other datasets like COCO and OpenImages. \textbf{Right: Summary of \Datasetname.} \Datasetname~is the
largest fully annotated vision dataset available to the general research
community, in terms of the total number of images, the number of concepts, and the number of bounding boxes (for object detection task).}
\label{Tab:summary_of_dataset}
\parbox{0.3\linewidth}{
\centering
\includegraphics[width=0.47\textwidth]{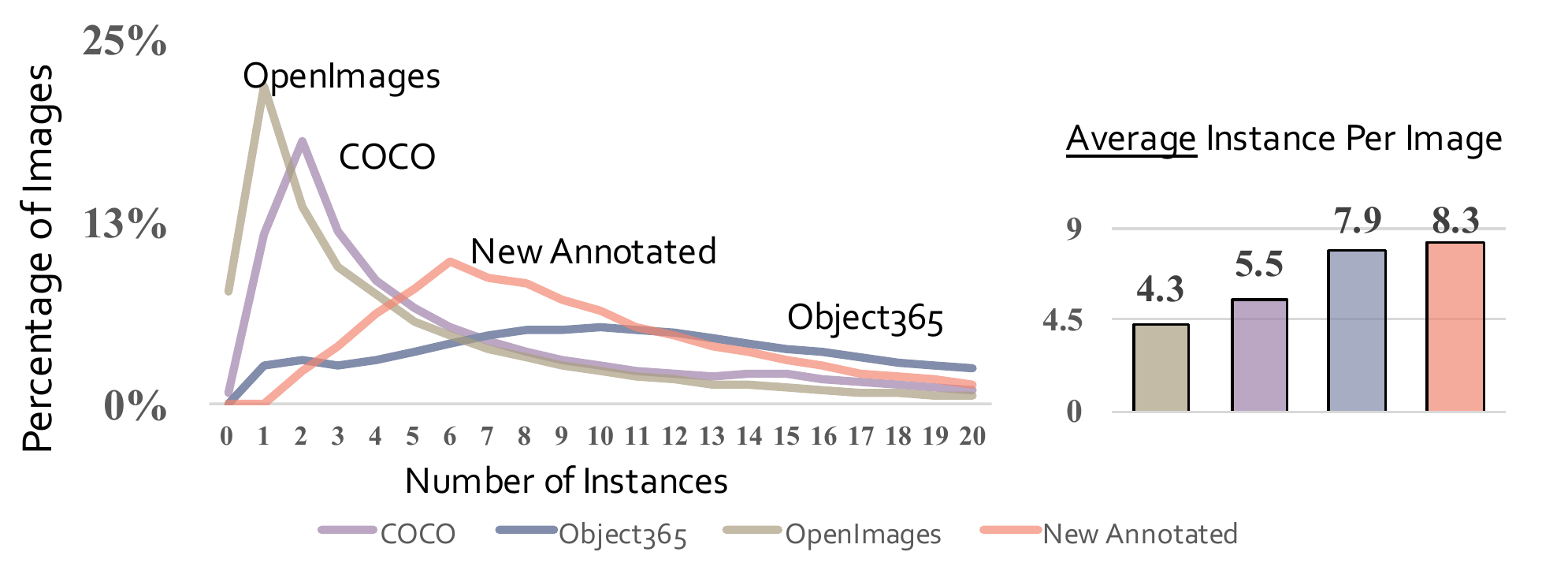}
}
\hfill
\parbox{0.85\linewidth}{
\centering
\scriptsize
\begin{tabular}{lllllll}
Datasets & Concepts & Images & Boxes  & Anno.\\ \Xhline{1.5pt}
YFCC-100M~\cite{thomee2016yfcc100m}   & -  & 100M   & -             & No        \\

ImageNet22K~\cite{deng2009imagenet}   & 22K  & 14M   & -             & Yes        \\

\textbf{Bamboo-CLS} & \textbf{119K} & \textbf{69M} & -  & Yes        \\ 
\Xhline{1pt}
COCO~\cite{COCO}    & 80 & 118K  & 1M  & Yes        \\
Objects365~\cite{shao2019objects365}  & 365 & 609K   & 10M & Yes        \\
OpenImages~\cite{kuznetsova2020open}    & 600 & 2M  & 14M & Partial        \\
\textbf{Bamboo-DET} & \textbf{809} & \textbf{3M} & \textbf{27M} & Yes        \\ 
\end{tabular}%
}
\end{table*}

As shown in Fig.~\ref{fig:statistics}, we illustrate the sorted distribution of image numbers per category in the Bamboo.
%
Generally, we emphasize that the new annotated data in the
Bamboo-CLS and Bamboo-DET are a powerful complement to the current public datasets---This data mostly belongs to tail classes of public datasets and new classes. In the following, we briefly describe the data statistics of Bamboo.  

\noindent \textbf{Image Classification (Bamboo-CLS).}
Bamboo-CLS~has 68,884,828 images spread across 119,035 categories.
42,648,217 out of 68,884,828 images are newly annotated, which is twice of ImageNet22K. In addition, 20,000 out of 119,035 categories are from Wikidata. These categories mainly are fine-grained concepts, such as \textit{Folland Midge} (one type of fighter) and \textit{hemaria hemishphaerica} (a species of fungi).
To our knowledge, Bamboo-CLS~is the
largest clean image dataset available to the vision research
community, in terms of the total number of images and categories.

\noindent \textbf{Object Detection (Bamboo-DET).}
Bamboo-DET has 3,104,012 images across 809 categories. 
Specifically, 557,457 images are newly annotated, and 150  concepts are from the Wikidata. In addition, we provide the statistics on instances per image of Bamboo-DET. As shown in Table~\ref{Tab:summary_of_dataset},  Our newly annotated data has 8.3 instances (on average) per image, which is dense than existing datasets, \ie MS-COCO, Object-365, and OpenImages.

%% file: sections/7.AlgorithmicAnalysis.tex
\section{Experiments}
\label{Algorithmic Analysis}

\subsection{Experimental Setups}

\subsubsection{Upstream Pre-training}
\noindent \textbf{Hyper-parameter.}
We train the models on 64 A100 GPUs for image classification, with a total batch size of 8,192. We employ an AdamW~\cite{loshchilov2017decoupled} optimizer of 30 epochs using a cosine decay scheduler with two epochs of linear warm-up. The ResNet-50 backbone is initialized from the model officially offered by PyTorch. The ViT B/16 backbone is initialized from ImageNet1K model provided by timm.\footnote{https://github.com/rwightman/pytorch-image-models/tree/master/timm} The weight decay, and warm-up learning rate are $2\times10^{-8}$, $10^{-6}$, and $2\times10^{-2}$.

\noindent \textbf{Datasets.}
Beyond the new annotated data, we include ImageNet22K~\cite{deng2009imagenet}, INaturalist2021~\cite{inaturalist}, Herbarium2021,\footnote{https://www.kaggle.com/c/herbarium-2021-fgvc8/overview}, Danish Fungi 2020~\cite{picek2021danish}, iWildCam2020~\cite{beery2021iwildcam}, TsinghuaDogs~\cite{Zou2020ThuDogs}, Places~\cite{places}, FoodX-251~\cite{kaur2019foodx}, CompCars~\cite{yang2015large} in the upstream classification dataset training.
We train the models on 48 A100 GPUs for detection, with a total batch size of 384, a total learning rate of $0.4$, SGD optimizer of momentum 0.9, and a weight decay of $0.0001$. We use the MultiStep learning rate scheduler with the decay rate of $0.1$ on [$16$, $22$] epochs and train for $26$ epochs in total. We also applied the warm-up learning rate of $0.0004$ for $1$ epoch. 
We used Cross-Entropy-Loss for categorization and Smoothed-L1-Loss for bounding box regression. 
Beyond the new annotated data, we include COCO~\cite{COCO}, Objects365~\cite{shao2019objects365} and OpenImages~\cite{kuznetsova2020open} in the upstream object detection dataset training.

\subsubsection{Downstream Evaluation}
\noindent \textbf{Datasets.}
In the following sections, we adopt the downstream datasets that are widely used in the transfer learning study~\cite{ImageNettransferbetter_1,VTAB,MOCOv1}. For models pre-trained on the image classification datasets, we use CIFAR10~\cite{cifar}, CIFAR100~\cite{cifar}, OxfordFlower~\cite{flowers}, Food101~\cite{food}, Caltech101~\cite{Caltech101}, OxfordPets~\cite{pets}, DTD~\cite{DTD}, StanfordCars~\cite{stanfordcar}, FGVC-Aircraft~\cite{FGVC}, SUN397~\cite{sun}, ImageNet1K~\cite{russakovsky2015imagenet} as the downstream evaluation datasets. As for the object detection task, we select PASCAL VOC~\cite{VOC} and CityPersons~\cite{CityPersons} as the downstream evaluation datasets. These datasets cover a wide range of image domains. The number of images in each dataset ranges from 2,000 to 80,000, and the number of classes in each dataset ranges from 10 to 8,000. 

\noindent \textbf{Evaluation Protocol.}
For the classification task, we use image features taken from the penultimate layer of each model, ignoring any classification layer provided. We train a logistic regression classifier for the linear probe evaluation setting.
We finetune the entire model loaded with its backbone and FPN weights for the detection task.
We only report the evaluation performance of models on downstream datasets.
We finetune the model on $8$ 1080-Ti GPUs for detection, with the batch size of $16$, SGD optimizer of momentum $0.9$, and weight decay $0.0001$ by loading the weights of backbone and FPN. We conduct a grid search on learning rate among [$5\times10^{-4}$, $1\times10^{-3}$, $5\times10^{-3}$, $1\times10^{-2}$]. The learning rate is decayed by 0.1 at 16 and 18 and stopped training at 19 epochs.

\begin{table*}[t]
\caption{\textbf{Downstream classification tasks performance among different pre-training methods.} 
Bamboo achieves the state-of-the-arts linear probe performance on the downstream tasks. Lang. indicates image-text pair. Numbers in \textcolor{myred}{red} are the performance gain on the same backbone network. \Datasetname~here refers to the Bamboo-CLS. Pets indicates OxfordPets. Flowers indicates OxfordFlower. Cars indicates StanfordCars. Aircraft indicates FGVC-Aircraft. IN1K indicates ImageNet1K. Results reported by the author are marked in \textcolor{gray}{gray}. We mainly compare with the methods conducted on supervised learning. Other performance of current methods are also presented.}
\centering
\ra{1.1}
\label{tab:summary_of_CLS}
\begin{tabular}{lllll|llllllllllll}
\rotatebox{90}{Method} & \rotatebox{90}{Data} & \rotatebox{90}{Annotation} & \rotatebox{90}{Model} & \rotatebox{90}{Paradigm}  & \rotatebox{90}{CIFAR10} & \rotatebox{90}{CIFAR100} & \rotatebox{90}{Food101}  & \rotatebox{90}{Pets}   & \rotatebox{90}{Flowers} & \rotatebox{90}{SUN397} & \rotatebox{90}{Cars} & \rotatebox{90}{DTD} &
\rotatebox{90}{Caltech101} &
\rotatebox{90}{Aircraft}  &\rotatebox{90}{IN1K} & \rotatebox{90}{AVG$\uparrow$}  \\
\Xhline{1.5pt}
SwAV~\cite{swav} & IN1K & 1.2M  & RN50 & Self. & 92.5    & 76.6     & 76.4    & 88.0 & 93.0    & 65.5   & 60.5 & 78.1 & 91.0 & 56.0 & 66.9 & 76.8 \\
DINO~\cite{DINO} & IN1K & 1.2M  & RN50 & Self. & 93.7	&79.2	&77.2	&89.2	&96.2	&66.0	&68.3 & 77.6 & 92.3	&63.1	& 83.3 &79.8 \\
SWSL~\cite{SWSL} & IG-1B & 1B & RN50 & Semi.& 94.7	&79.5	&79.1	&94.4	&94.6	&67.8	&65.9 & 77.8 & 96.1	&58.4 & 81.2 &80.9 \\
WSL~\cite{IN-1B} & IG-1B & 1B & RX101 & Weak. &95.0    & 78.2     & 83.5    & \textbf{95.5} & 90.8    & 67.9   & 72.3 & 75.3 & 93.3 & 53.9  & 83.3 & 81.0 \\ 
\color{gray}{CLIP}~\cite{CLIP} & \color{gray}{WIT}  & \color{gray}{400M} & \color{gray}{RN50} & \color{gray}{Lang.} &\color{gray}{88.7}	&\color{gray}{70.3}	&\color{gray}{86.4}	&\color{gray}{88.2}	&\color{gray}{96.1}	&\color{gray}{73.3}	&\color{gray}{78.3}
&\color{gray}{76.4} &\color{gray}{89.6}
&\color{gray}{49.1} & \color{gray}{73.3} & \color{gray}{79.1} \\

\color{gray}{CLIP}~\cite{CLIP} & \color{gray}{WIT}  & \color{gray}{400M} & \color{gray}{B/16} & \color{gray}{Lang.} &\color{gray}{96.2}	&\color{gray}{83.1}	&\color{gray}{92.8}	&\color{gray}{93.1}	&\color{gray}{98.1}	&\color{gray}{78.4}	&\color{gray}{86.7}
&\color{gray}{79.2}	&\color{gray}{94.7}
&\color{gray}{59.5} & \color{gray}{80.2} & \color{gray}{85.6} \\
\Xhline{1pt}
BiT~\cite{bit} & IN1K & 1.2M & RN50 & Sup.   & 91.7	&74.8	&72.5	&92.3	&92.0	&61.1	&53.5 & 72.4 & 91.2 	&52.5	&75.2 &73.6\\
BiT~\cite{bit} & IN22K  & 14M & RN50 & Sup. & 94.9	&82.2	&83.3	&91.5	&99.4	&69.9	&59.0 & 77.3 & 93.9	&55.6 &76.7 &80.3\\

\Xhline{1pt}
RN50 & \Datasetname  & 69M & RN50 & Sup. & 93.9	& 81.2	& 85.3	& 92.0	& 99.4	&72.2	& 91.1 & 76.5 & 93.2	& 84.0	& 77.2 & 86.0 \textbf{\textcolor{myred}{(+5.1)}}
 \\
B/16 & \Datasetname  & 69M & B/16 & Sup. & \textbf{98.2}	& \textbf{90.2}	& \textbf{92.9}	& 95.1	& \textbf{99.8}	& \textbf{79.0}	& \textbf{93.3}	& \textbf{81.2} & \textbf{97.0} & \textbf{88.1} & \textbf{83.6} & \textbf{91.8} \textbf{\textcolor{myred}{(+6.2)}}
 \\
\end{tabular}
\end{table*}

\begin{table}[t]
\scriptsize
\caption{\textbf{Comparisons of downstream detection tasks performance.} 
Pre-trained model on \Datasetname~achieves significant performance gain. \Datasetname~here refers to the Bamboo-DET. VOC means the PASCAL VOC dataset~\cite{VOC}. CITY. means the CityPersons dataset~\cite{CityPersons}.}
\label{Tab:results_bamboo_det}
\centering
\footnotesize
\begin{tabular}{ll|lll}
 \multirow{2}{*}{Data} & \multirow{2}{*}{Anno.}   & VOC & CITY. & COCO   \\
     &  & AP50 $\uparrow$ & MR $\downarrow$ & mmAP $\uparrow$    \\
\Xhline{1.5pt}
\color{gray}{COCO}~\cite{shao2019objects365} & \color{gray}{1M}    & \color{gray}{85.1} & \color{gray}{16.2} & -   \\
\color{gray}{OpenImages}~\cite{shao2019objects365} & \color{gray}{14M}   & \color{gray}{82.4} & \color{gray}{16.8} & \color{gray}{37.4} \\
Objects365 &  10M  & 86.4 & 14.7 & 39.3\\
\Xhline{1pt}
\Datasetname  & 27M  & \textbf{87.5} \textbf{\textcolor{myred}{(+1.1)}} & \textbf{12.6} \textbf{\textcolor{myred}{(+2.1)}} & \textbf{43.9} \textbf{\textcolor{myred}{(+4.4)}}\\
\end{tabular}%
\end{table}

\begin{table*}[t]
\caption{\textbf{Comparisons of zero-shot downstream classification tasks performance among different pre-training methods.} 
Bamboo achieves the state-of-the-arts linear probe performance on the downstream tasks. Lang. indicates image-text pair. Numbers in \textcolor{myred}{red} are the performance gain on the same backbone network. \Datasetname~here refers to the Bamboo-CLS. Pets indicates OxfordPets. Flowers indicates OxfordFlower. Cars indicates StanfordCars. Aircraft indicates FGVC-Aircraft. IN1K indicates ImageNet1K. Results reported by the author are marked in \textcolor{gray}{gray}. We mainly compare with the methods conducted on supervised learning. Other performance of current methods are also presented.}
\centering
\ra{1.1}
\label{tab:zero-shot}
\begin{tabular}{lllll|llllllllllll}
\rotatebox{90}{Method} & \rotatebox{90}{Data} & \rotatebox{90}{Annotation} & \rotatebox{90}{Model} & \rotatebox{90}{Paradigm}  & \rotatebox{90}{CIFAR10} & \rotatebox{90}{CIFAR100} & \rotatebox{90}{Food101}  & \rotatebox{90}{Pets}   & \rotatebox{90}{Flowers} & \rotatebox{90}{SUN397} & \rotatebox{90}{Cars} & \rotatebox{90}{DTD} &
\rotatebox{90}{Caltech101} &
\rotatebox{90}{Aircraft}  &\rotatebox{90}{IN1K} & \rotatebox{90}{AVG$\uparrow$}  \\
\Xhline{1.5pt}
\color{gray}{CLIP}~\cite{CLIP} & \color{gray}{WIT}  & \color{gray}{400M} & \color{gray}{RN50} & \color{gray}{Lang.} &\color{gray}{{91.6}}	&\color{gray}{{68.7}}	&\color{gray}{{89.2}}	&\color{gray}{{88.9}}	&\color{gray}{70.4}	&\color{gray}{{65.2}}	&\color{gray}{{65.6}}
&\color{gray}{46}	&\color{gray}{{89.3}}
&\color{gray}{27.1} & \color{gray}{68.6} & \color{gray}{70.0} \\
RN50 & \Datasetname  & 69M & RN50 & Sup. & 93.8	& 67.7	& 81.6	& 74.3	& {87.3}	&58.7	& 63.0 & {51.1} & 88.4	& {87.2}	& {82.5} & {76.0} \textbf{{\textcolor{myred}{(+6.0)}}}
 \\
\end{tabular}
\end{table*}

\begin{table*}[t]
\caption{\textbf{Comparisons of fine-tuning downstream classification tasks performance among different pre-training methods.} 
Bamboo achieves the state-of-the-arts fine-tuning performance on the downstream tasks.}
\centering
\ra{1.1}
\label{tab:fine-tuning}
\begin{tabular}{lllll|llllllllllll}
\rotatebox{90}{Method} & \rotatebox{90}{Data} & \rotatebox{90}{Annotation} & \rotatebox{90}{Model} & \rotatebox{90}{Paradigm}  & \rotatebox{90}{CIFAR10} & \rotatebox{90}{CIFAR100} & \rotatebox{90}{Food101}  & \rotatebox{90}{Pets}   & \rotatebox{90}{Flowers} & \rotatebox{90}{SUN397} & \rotatebox{90}{Cars} & \rotatebox{90}{DTD} &
\rotatebox{90}{Caltech101} &
\rotatebox{90}{Aircraft}  &\rotatebox{90}{IN1K} & \rotatebox{90}{AVG$\uparrow$}  \\
\Xhline{1.5pt}
DINO & IN1K & 1.2M & RN50 & Self. & 97.1 & 84.0 & 86.3 & 90.0 & 96.1 & 65.2 & 84.6 & 77.6 & 91.4 & 81.8 & 66.5 & 83.7 \\
SWAV & IN1K & 1.2M & RN50 & Self. & 97.2 & 84.2 & 86.0 & 90.3 & 95.7 & 64.4 & 83.9 & 77.2 & 91.7 & 81.2 & 66.9 & 83.5 \\
SWSL & IG-1B & 1B & RN50 & Semi. & 97.0 & 86.5 & 87.3 & 94.4 & 97.0 & 66.0 & 88.5 & 78.3 & 93.8 & 84.0 & 81.7 & 86.8 \\
BiT-S & IN1K & 1.2M & RN50 & Sup. & 97.0 & 85.0 & 85.7 & 92.8 & 95.0 & 60.3 & 87.5 & 74.7 & 92.0 & 83.8 & 75.2 & 84.5 \\
BiT-M & IN22K & 14M & RN50 & Sup. & 97.6 & 86.2 & 87.9 & 91.5 & 98.1 & 64.2 & 88.2 & 78.4 & 92.9 & 84.3 & 76.7 & 86.0 \\
\Xhline{1pt}
RN50  & Bamboo & 69M  & RN50 & Sup.  & 97.3 & 87.0 & 87.5 & 92.0 & 99.4 & 72.2 & 91.4 & 77.1 & 93.9 & 85.9 & 77.1 & 87.3 \textbf{\textcolor{myred}{(+0.5)}}
 \\
\end{tabular}
\end{table*} 

\subsection{Power of \Datasetname~as Pre-Training}
\subsubsection{Main Results}

\noindent \textbf{Information-Dense Annotations Matter.} As shown in Table~\ref{tab:summary_of_CLS}, ResNet-50 (RN50) pre-trained on CLIP (400M) or IG-1B (1B) achieves better downstream task performance than BiT pre-trained on ImageNet1K (IN1K)~\cite{russakovsky2015imagenet}. However, compared to RN50 pre-trained on \Datasetname,
CLIP-RN50 or RN50 pre-trained on IG-1B achieves inferior performance. 

It indicates that the amount of informative-dense annotations instead of the sheer number of annotations is much more essential for model pre-training.
Compared to CLIP, which leverages the vast amount of image-text pairs on the web for pre-training, our \Datasetname~presents an active and continual framework that collects and annotates fully-supervised samples in a highly scalable manner. 

\noindent \textbf{Comprehensive Label System Helps.}
As shown in Table~\ref{tab:summary_of_CLS}, most methods pre-trained on IN1K, IG-1B, or WIT achieve more than 90\% accuracy on the OxfordPets and OxfordFlower. But they only achieve less than 80\% accuracy on the StanfordCars and FGVC-Aircraft. It indicates that these pre-trained datasets might include more semantic concepts related to OxfordPets and OxfordFlower.
Our \LN~spreads a large spectrum of concepts. Notably, it includes much more concepts that are neglected in the current public and nonpublic datasets.
As a result, models pre-trained on \Datasetname~achieve much better performance than other methods.
Beyond general object detection, it is also
important to validate the generalization ability on specific object detection problems like pedestrian detection. 
%
 
\noindent \textbf{Bamboo is an Effective Pre-Training Source.} Compared to other methods, \Datasetname~achieves the best performance among downstream tasks on average. 
As shown in Table~\ref{tab:summary_of_CLS}, ViT B/16 pretrained on \Datasetname~outperforms CLIP with 6.2 points gain. 
It indicates that our annotation is much more informative and hence more helpful for the model pre-training.
In addition, Table~\ref{Tab:results_bamboo_det} presents that ResNet-50 with FPN pretrained on \Datasetname~outperforms Objects365 with 1.1 points gain on PASCAL VOC and 2.1 points gain on CityPersons. 

\subsubsection{Further Analysis}

\noindent \textbf{The Influence Of Similar Semantic Proposals.}
The total annotation cost for the object detection task depends on the number of proposals. Images with dense proposals are more expensive than sparse ones. Based on our observation, many proposals with similar semantics tend to form a group in a single image. To evaluate their effectiveness, we conduct the following experiments on Objects365~\cite{shao2019objects365} dataset.

Firstly, we define an image as a crowded image if it contains at least one category with more than $15$ proposals. By removing all 27K crowded images from the full Objects365 dataset, we denote the remaining part as Objects365-sparse. Keeping the number of proposals the same as Objects365-sparse, we randomly removed 90K images from the full Objects365 dataset and marked the remaining part as Objects365-random. Furthermore, keeping the total object amount the same as Objects365-sparse, we randomly removed 101K non-crowded images from the full Objects365 dataset and denoted the remaining part as Objects365-dense.

\begin{center}
\vspace{2pt}
\begin{tabular}{lll|l}
\multirow{2}{*}{\small{Dataset}} & \multirow{2}{*}{\small{Images}} & \multirow{2}{*}{\small{Proposals}} & \small{VOC} \\
&   &  & \small{AP50} \small{$\uparrow$} \\
\Xhline{1.5pt}
\small{Objects365-sparse} & \small{581K} & \small{8.2M} & \small{86.3} \\
\small{Objects365-random} & \small{519K} & \small{8.2M} & \small{85.8} \\
\small{Objects365-dense} & \small{508K} & \small{8.2M} & \small{85.1} \\
\end{tabular}
\vspace{5pt}
\end{center}

Given the same annotation budget, we find that choosing to label non-crowded images yields better results for pre-training performance. Therefore, as mentioned in Sec.~\textcolor{red}{3.3.2} of the main paper, we filter out covariate shift data in the OOD rectification step.

\noindent \textbf{Finetuning Transfer.}
We compared our model pre-trained on \Datasetname~to various with the ResNet-50 backbone. We present the finetuning transfer performance of the models pre-trained on \Datasetname. The finetuning strategy among each downstream task is followed by the SimCLR~\cite{simclr}.
Table~\ref{tab:fine-tuning} shows the comparison. \Datasetname~model achieves a 1.3\% average accuracy gain compared to BiT-M pre-trained on the current largest public classification dataset: ImageNet22K. It indicates a larger, carefully annotated dataset can continually improve the performance of models.
Besides, \Datasetname~model achieves a 0.5\% average accuracy gain compared to SWSL, pre-trained on the IG-1B with 1B weakly supervised hashtags. \Datasetname~is 20 times smaller than IG-1B, which indicates that the amount of informative-dense annotations instead of the sheer number of weak annotations is much more essential for model pre-training.

\noindent \textbf{Zero-Shot Transfer.}
We present the zero-shot transfer performance of the models pre-trained on \Datasetname. We compared our model pre-trained on \Datasetname~to CLIP models with the same backbone. 

Table~\ref{tab:zero-shot} shows the comparison.  We can indicate that \Datasetname~model conclusively outperforms CLIP model with the same backbone: RN50. Specifically, \Datasetname~model achieves a 6\% average accuracy gain. On the FGVC-Aircraft, \Datasetname~model achieves 87.2\%, despite having never seen any training images from this dataset. 
\Datasetname~includes all the concepts in the downstream tasks. However, we conduct data overlap analysis of \Datasetname~in Sec.~\ref{sec:Data_Overlap_Analysis}, ensuring \Datasetname~rarely includes downstream data.

\noindent \textbf{Robustness to Natural Distribution Shift.}
We conduct experiments on the ObjectNet~\cite{barbu2019objectnet} to compare \Datasetname~models with other models when evaluated on the data with controls for rotation, background, and viewpoint. ObjectNet is a dataset collected in the real world, where multiple objects
are always present.
There are 313 object classes in
total, with 113 overlapping with ImageNet1K. We follow the literature~\cite{bit,CLIP} and
evaluate our models on those 113 classes.

\vspace{3pt}
\begin{center}
\begin{tabular}{cccc|c}
\footnotesize {Method} & \footnotesize{Data} & \footnotesize{Model} & \footnotesize{Para.}  & \footnotesize{ObjectNet $\uparrow$}  \\
\Xhline{1.5pt}
\footnotesize{\color{gray}{BiT-L}~\cite{bit}} & \footnotesize{\color{gray}{JFT-300M}} &  \footnotesize{\color{gray}{RN50}} & \footnotesize{\color{gray}{Weak.}} & \footnotesize{\color{gray}{37.6}} \\
\footnotesize{\color{gray}{ANN-1.3B}~\cite{beal2021billion}} & \footnotesize{\color{gray}{ANN-1.3B}} & \footnotesize{\color{gray}{B/16}} & \footnotesize{\color{gray}{Weak.}} & \footnotesize{\color{gray}{50.7}} \\
\Xhline{1pt}
\footnotesize{RN50} & \footnotesize{Bamboo} & \footnotesize{RN50} & \footnotesize{Sup.} & \footnotesize{\textbf{38.8} \textbf{\textcolor{myred}{(+1.2)}}} \\
\footnotesize{B/16} & \footnotesize{Bamboo} & \footnotesize{B/16} & \footnotesize{Sup.} & \footnotesize{\textbf{53.9} \textbf{\textcolor{myred}{(+3.2)}}} \\
\end{tabular}
\end{center}
\vspace{5pt}

As shown above, we compare \Datasetname~models with the state-of-the-art model with the same backbone. Specifically, ResNet-50 pre-trained on \Datasetname~achieves 1.2\% gains compared with ResNet-50 pre-trained on JFT-300M.  
ViT B/16 pre-trained on \Datasetname~achieves 3.2\% gains compared with ViT B/16 pre-trained on Anno-1.3B.
Even though JFT-300M and Anno-1.3B are much larger than \Datasetname, the informative data in \Datasetname~is more helpful for pre-trained models in real scenarios.

%% file: sections/10.social_impact.tex
\section{Social Impact}
\label{sec:Data_Overlap_Analysis}
The proposed \Datasetname~dataset and pre-training model shows the capacity and generalization of learned image representation which
could benefit many applications of computer vision. 
However, our data usage might bring several risks, such as data overlapping, privacy, and inappropriate content. We
discuss these risks and their mitigation strategies as follows.

\noindent \textbf{Data Overlapping.} A concern with pre-training on an extensive dataset
is unintentional overlap with downstream evaluation~\cite{CLIP}. To enable a meaningful test of generalization, we identify and remove all duplicates among upstream data. 
Specifically, we utilize Difference Hash (DHash)~\cite{DHash} to present the information of each image.
We calculate the hash-code of each downstream image and each crawled image, and two images with the same hash-code are regarded as similar ones. Then, we filter out the crawled images that are similar to downstream images.
Based on the above method, we discard 122,939 images for classification and 1,046 images for detection from the unlabeled pool. 

\noindent \textbf{Copyright.} We crawl only the data under the Creative Commons license (CC-BY) for the Bamboo-DET. This license allows free use, redistribution, and adaptation for non-commercial purposes.  
For the Bamboo-CLS data, 30\% of data is under the CC-BY license because of its large volume of data.
For Bamboo-CLS data that is not under the CC-BY license, referred to LAION-400M~\cite{laion400m} and Conceptual 12M~\cite{cc12m}, we only present the lists of URLs to this data without redistribution. We build the meta file as follow.
\begin{equation}
\text{[image\_url]\ [class\_index]} 
\nonumber
\end{equation}

Referred to \textit{Authors Guild, Inc. v. Google Inc.}~\cite{googlelaw}, training data on the copyrighted works might be considered as transformative uses and was thus might be regarded as \textit{Fair Use}\footnote{\label{fair_use}https://www.copyright.gov/fair-use/index.html}. In addition, referred to \textit{Article 30-4 of the new Copyright Act}~\cite{Copyrightjapan}, there are no restrictions on the subject, purpose, and method of data analysis, and there is no obligation to compensate the copyright holder. 
However, we admit that using copyright material as training data is still a controversial issue in Artificial Intelligence, and we would no doubt follow the newest law worldwide. 
Bamboo is specifically open for non-commercial research and/or educational purposes to respect the copyright law. For researchers and educators who wish to use copyrighted images for that purpose, training or benchmarking models with copyrighted works would be qualified as \textit{transformative} uses and thus not infringe copyright law in the U.S.\ref{fair_use}. Nevertheless, the users must strictly follow the Flickr Terms of Use.\footnote{https://www.flickr.com/help/terms/api} And the users of these images accept full responsibility for the use of the image. 

\noindent \textbf{Problematic Content.} The inappropriate contents such as drugs, nudity, and other offensive content exist in the web data. we ask annotators to discard such images instead of conducting annotation.

\noindent \textbf{Privacy.} To mitigate privacy issues with public visual
datasets, researchers have attempted to obfuscate private information before publishing the data~\cite{privacy1,privacy2}. We plan to follow this line of work to blur faces, and license plates in our new annotated data.
In addition, if the original picture found at the URL present on the Bamboo on the record states users' names, phone numbers, or any personal information, users can request a takedown of this image.

\noindent \textbf{Bias.} The images were crawled from Flickr, thus inheriting all the biases of that website. The usage of user-generated data might bring the risk of bias. We plan to tackle this problem by balancing various categories.

%% file: sections/8.Discussion.tex
\section{Conclusion}
\label{Conclusion}
In our work, with a human-machine synergy, we actively and continually build a mega-scale and information-dense dataset, namely \Datasetname. \Datasetname~is the
largest clean image dataset available to the vision research
community, in terms of the total number of images and the number of categories, for classification and detection tasks.
Our key insight is that a unified and visually-oriented label system is crucial for model pre-training, and rectifying OOD samples is indispensable for AL to function in realistic scenarios.
We have demonstrated the effectiveness of \Datasetname~as a better pre-training dataset for various downstream tasks and provided several valuable observations. 